\documentclass[10pt,twocolumn,letterpaper]{article}

\usepackage{cvpr}              %

\usepackage[accsupp]{axessibility} %

\def\cvprPaperID{3073} %
\def\confName{CVPR}
\def\confYear{2022}

\usepackage{array}
\newcolumntype{P}[1]{>{\centering\arraybackslash}p{#1}}

\usepackage[table,dvipsnames]{xcolor}
\usepackage{times}
\usepackage{latexsym}

\usepackage{microtype}

\usepackage{graphicx}               %
\usepackage{tabularx}               %
\newcolumntype{C}{>{\centering\arraybackslash}X}
\usepackage{multirow}               %
\usepackage{diagbox}                %
\usepackage{hhline}                 %
\usepackage{color}                  %
\usepackage{amsmath}                %
\usepackage{amssymb}                %
\usepackage{mathtools}              %
\usepackage{enumitem}               %
\newcommand{\bs}{\boldsymbol}

\newcommand{\real}{\mathbb{R}}

\usepackage{booktabs}
\usepackage{extarrows}
\usepackage{makecell}
\usepackage{dsfont}

\newtheorem{thm:def}{Definition}
\newtheorem{thm:eg}{Example}
\newtheorem{thm:lem}{Lemma}

\newcommand{\mmee}{M\textsuperscript{2}E\textsuperscript{2} }

\DeclareMathOperator*{\argmax}{arg\,max}

\usepackage[pagebackref,breaklinks,colorlinks]{hyperref}

\usepackage[capitalize]{cleveref}
\crefname{section}{Sec.}{Secs.}
\Crefname{section}{Section}{Sections}
\Crefname{table}{Table}{Tables}
\crefname{table}{Tab.}{Tabs.}

\definecolor{RoseQuartzBg}{HTML}{F7CAC9}
\definecolor{RoseQuartz}{HTML}{F5A798}
\definecolor{Serenity}{HTML}{92A8D1}
\definecolor{OrangeRed}{rgb}{1.0, 0.27, 0.0}
\definecolor{Turquoise}{HTML}{0F4C81}
\definecolor{mint}{rgb}{0.24, 0.71, 0.54}
\definecolor{green}{rgb}{0.0, 0.120, 0.0}

\usepackage{xparse}
\NewDocumentCommand{\heng}{ mO{} }{\textcolor{OrangeRed}{\textsuperscript{\textit{Heng}}\textsf{\textbf{\small[#1]}}}}
\NewDocumentCommand{\ruochen}{ mO{} }{\textcolor{Serenity}{\textsuperscript{\textit{Ruochen}}\textsf{\textbf{\small[#1]}}}}
\NewDocumentCommand{\shuohang}{ mO{} }{\textcolor{red}{\textsuperscript{\textit{Shuohang}}\textsf{\textbf{\small[#1]}}}}
\NewDocumentCommand{\luowei}{ mO{} }{\textcolor{TealBlue}{\textsuperscript{\textit{Luowei}}\textsf{\textbf{\small[#1]}}}}
\NewDocumentCommand{\xudong}{ mO{} }{\textcolor{mint}{\textsuperscript{\textit{Xudong}}\textsf{\textbf{\small[#1]}}}}
\NewDocumentCommand{\manling}{ mO{} }{\textcolor{Turquoise}{\textsuperscript{\textit{Manling}}\textsf{\textbf{\small[#1]}}}}
\NewDocumentCommand{\cz}{ mO{} }{\textcolor{red}{\textsuperscript{\textit{Chenguang}}\textsf{\textbf{\small[#1]}}}}
\NewDocumentCommand{\shihfu}{ mO{} }{\textcolor{purple}{\textsuperscript{\textit{Shif-Fu}}\textsf{\textbf{\small[#1]}}}}
\NewDocumentCommand{\michael}{ mO{} }{\textcolor{blue}{\textsuperscript{\textit{Michael}}\textsf{\textbf{\small[#1]}}}}
\NewDocumentCommand{\yi}{ mO{} }{\textcolor{olive}{\textsuperscript{\textit{Yi}}\textsf{\textbf{\small[#1]}}}}

\begin{document}

\title{CLIP-Event: Connecting Text and Images with Event Structures}

\author{
\textbf{
Manling Li\textsuperscript{\textnormal{1}}\thanks{The work is done when the first author was an intern at Microsoft. }, 
Ruochen Xu\textsuperscript{\textnormal{2}}, Shuohang Wang\textsuperscript{\textnormal{2}}, Luowei Zhou\textsuperscript{\textnormal{2}}, 
Xudong Lin\textsuperscript{\textnormal{3}} 
}
\\
 \textbf{ Chenguang Zhu\textsuperscript{\textnormal{2}}, Michael Zeng\textsuperscript{\textnormal{2}}, Heng Ji\textsuperscript{\textnormal{1}}, Shih-Fu Chang\textsuperscript{\textnormal{3}}} 
 \\
  \textsuperscript{1}University of Illinois Urbana-Champaign  \ \ 
  \textsuperscript{2}Microsoft Research \ \ 
  \textsuperscript{3}Columbia University\\
  \texttt{\fontfamily{pcr}\selectfont\{manling2,hengji\}@illinois.edu},\\
  \texttt{\fontfamily{pcr}\selectfont{\{ruox,shuowa,luozhou,chezhu,nzeng\}@microsoft.com}}\\
  \texttt{\fontfamily{pcr}\selectfont{\{xudong.lin,sc250\}@columbia.edu}}\\
  }
  
\maketitle

\date{}

\maketitle

\begin{abstract}
Vision-language (V+L) pretraining models have achieved great success in supporting multimedia applications by understanding the alignments between images and text. 
While existing vision-language pretraining models primarily focus on  understanding  objects in images or entities in text, they often ignore the alignment at the level of events and their argument structures. %
In this work, we propose a contrastive learning framework to enforce vision-language pretraining models to comprehend events and associated argument (participant) roles.
To achieve this, we take advantage of text information extraction technologies to obtain event structural knowledge, and utilize multiple prompt functions to contrast difficult negative descriptions by manipulating event structures. We also design an event graph alignment loss based on optimal transport to capture event argument structures. 
In addition, we collect a large event-rich dataset (106,875 images) for pretraining, which provides a more challenging image retrieval benchmark to assess the understanding of complicated lengthy sentences\footnote{The data and code are publicly available for research purpose in \url{https://github.com/limanling/clip-event}.}.
Experiments show that our zero-shot CLIP-Event outperforms the state-of-the-art supervised model in argument extraction on Multimedia Event Extraction, achieving more than 5\% absolute F-score gain in event extraction, as well as significant improvements on a variety of downstream tasks under zero-shot settings.

\end{abstract}

\begin{figure}[!h]
  \centering
   \includegraphics[width=1.0\linewidth,height=10em]{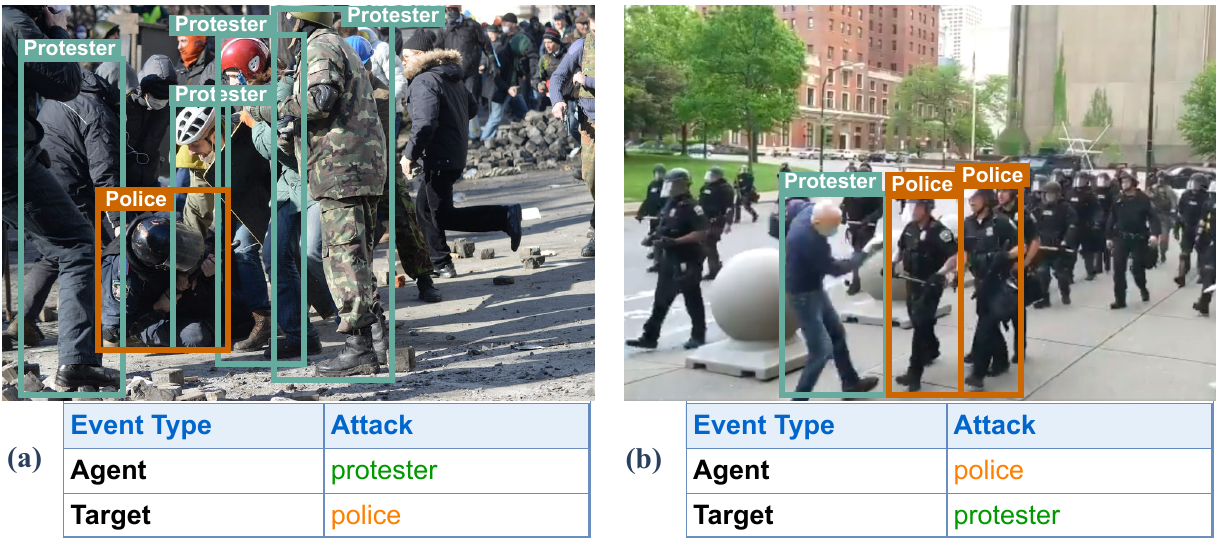}
   \caption{Examples of visual event \textsc{Attack} with different arguments. Groundings are bounding-boxes colored to match roles. }
   \label{fig:intro-example}
\end{figure}

\begin{figure*}[ht]
  \centering
   \includegraphics[width=1.0\linewidth]{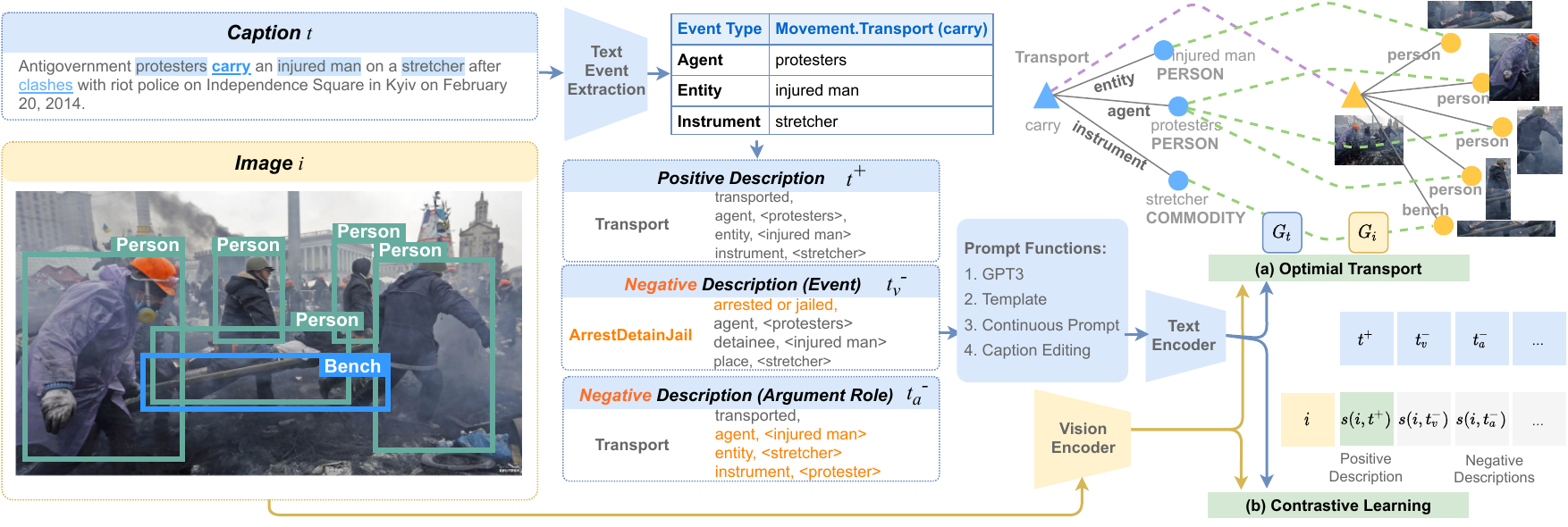}
   \caption{Architecture of CLIP-Event. We take advantage of event structural knowledge in captions to contrast hard negatives about event types and argument roles (in \textcolor{RoyalBlue}{blue}), which is then used to supervise image event understanding (in \textcolor{YellowOrange}{yellow}) as a cross-media transfer of event knowledge. 
   The negative event structures are highlighted in \textcolor{Orange}{orange}.
The events and objects are from automatic system output. 
} 
   \label{fig:framework}
\end{figure*}

\section{Introduction}

Real-world multimedia applications require an understanding of not only entity knowledge (i.e., objects and object types), but also event knowledge (i.e., event types) with event argument structures (i.e., entities involved and their roles). %
For example, 89\% images include events in contemporary multimedia news data\footnote{We randomly check 100 images at {\tiny \url{https://www.voanews.com/}}.}. %
Furthermore, recognizing the arguments (participants) is critical for news comprehension, since events might be contradictory if the arguments play different roles. %
For example, both \cref{fig:intro-example}(a) and \cref{fig:intro-example}(b) are about the same event type \textsc{Attack} and contain entities  \textit{protester} and \textit{police}, but their argument roles are different, i.e., the \textit{protester} plays the role of \textsc{attacker} in the first event and the role of \textsc{target} in the second event, and vice versa for the \textit{police}. Different argument roles for the same group entity result in the differentiation of two {attack} events. %

However, existing vision-language pretraining models~\cite{tan2019lxmert, chen2020uniter, li2020oscar, zhang2021vinvl, radford2021learning,jia2021scaling} focus on the understanding of images or entities, ignoring the event semantics and structures. As a result,  apparent failures happen in the circumstances requiring verb comprehension~\cite{hendricks2021probing}. 
Thus, we focus on integrating event structural knowledge into vision-language pretraining.  
Previous work primarily represents visual events as verbs with subjects and objects~\cite{sung2012unstructured,yao2010modeling,kato2018compositional,li2019transferable,wang2020learning,zhou2020cascaded}. 
However, events contain structural knowledge, with 
each event being assigned to an \textit{event type} that represents a set of synonymous verbs. Each argument is grounded to text or images, and associated with an \textit{argument role} that the participant is playing. As shown in \cref{fig:framework}, the \textit{carry} event is typed as \textsc{Transport}, with \textit{protesters} as \textsc{Agent}, \textit{injured man} as \textsc{Entity} and \textit{stretcher} as \textsc{Instrument}. %

There has been little research~\cite{ pratt2020grounded, li2020cross} on extracting event structures from news images, yielding limited support for event knowledge acquisition needed in downstream applications.  %
Thus, we propose to leverage text information extraction technologies, which have been well researched in natural language processing, to automatically extract event structures from captions. The captions essentially refer to the same event as the images in news data, e.g., 87\% captions describe the events in  the images\footnote{The statistics are on those mentioned above 100 images from VOA~\cite{voa}.}. Therefore, we design a self-supervised contrastive learning framework, \textbf{CLIP-Event}, using the rich event knowledge in captions as distant supervision to interpret events in the associated images, to effectively transfer event knowledge across modalities. %

In addition, in order to train robust representations capable of discriminating subtle differences between event types (e.g. \textsc{Transport} and \textsc{Arrest}) and argument roles (e.g. \textsc{Attacker} and \textsc{Victim}) using only images, we propose to generate \textit{hard negatives} by manipulating event structures, 
We translate both correct and manipulated event structures into text descriptions using an extensive set of \textit{event prompt functions}.
Following the state-of-the-art vision-language pretraining model CLIP~\cite{radford2021learning}, we optimize a contrastive learning objective between images and event-aware text descriptions.

Furthermore, to transfer knowledge of argument structures, we explicitly construct event graphs consisting of event types and argument roles in vision and text. 
We introduce a \textit{fine-grained} alignment between two event graphs, aligning the objects in images with the corresponding text entities and their argument roles. %
We employ optimal transport to encourage a \textit{global} alignment based on the structures of two graphs, which enables the model to capture the interactions between arguments. %
For example,  objects with similar visual features  tend to be aligned to the same argument role.

Our evaluations mainly focus on zero-shot settings, since it is crucial to understand new or previously unidentified events in real-world applications. %
Traditional methods based on limited pre-defined event ontologies are inapplicable in dealing with open-world events. %
Our pretrained model, on the other hand, is able to identify event structure using the natural language description of any unseen type and argument role, enabling zero-shot multimedia event extraction.

The evaluations on Multimedia Event Extraction~\cite{li2020cross} and Grounded Situation Recognition~\cite{pratt2020grounded} show that CLIP-Event significantly outperforms state-of-the-art vision-language pretraining models,  under both zero-shot settings and supervised settings. %
Furthermore, it 
and achieves significant gains in various downstream tasks under zero-shot settings such as image retrieval~\cite{datta2008image}, visual commonsense reasoning~\cite{zellers2019recognition} and visual commonsense reasoning in time~\cite{park2020visualcomet}.

In summary, this paper makes the following contributions: 
\begin{itemize}%
\item We are the first to exploit the visual event and argument structure information in vision-language pretraining.
\item {We introduce a novel %
framework by contrasting with negative event  descriptions, which are generated by various prompt functions conditioned on hard negative events and arguments. }
\item We propose event graph alignment based on optimal transport, extending previous image or object alignment to event structure aware alignment. %
\item We release an event-rich image-caption dataset with  106,875 images, including the extracted event knowledge, which can serve as a challenging image retrieval benchmark for evaluating the ability to understand complex and lengthy sentences in real-world applications. %
\end{itemize}

\section{Our Approach}

\begin{table*}[!hbt]
\footnotesize
\centering
\setlength\tabcolsep{2pt}
\setlength\extrarowheight{1pt}
\begin{tabular}{p{0.08\linewidth} p{0.10\linewidth} p{0.80\linewidth} } %
\toprule 
\textbf{Prompt} & \multicolumn{2}{c}{\textbf{Example descriptions of \cref{fig:framework} with \textcolor{OrangeRed}{arrest} as negative event} }
\\
\midrule

\multirow{5}{*}{\textbf{\makecell[l]{Single\\ Template}} } & 
\textbf{Template} & 
$\langle$arg1$\rangle$ \textbf{transported} $\langle$arg2$\rangle$ \textcolor{RoyalBlue}{in} $\langle$arg3$\rangle$ \textcolor{RoyalBlue}{instrument from} $\langle$arg4$\rangle$ \textcolor{RoyalBlue}{place to} $\langle$arg5$\rangle$ \textcolor{RoyalBlue}{place.}
\\ \cmidrule{2-3}
&
\textbf{Positive} & {\underline{\textit{Protesters}}} \textbf{transported} {\underline{\textit{an injured man}}} \textcolor{RoyalBlue}{in} {\underline{\textit{a stretcher}}} \textcolor{RoyalBlue}{instrument}. %
\\ \cmidrule{2-3}
&
\textbf{Negative-Evt} &
{\underline{\textit{Protesters}}} \textcolor{OrangeRed}{\textbf{arrested}} {\underline{\textit{an injured man}}} \textcolor{RoyalBlue}{in} {\underline{\textit{a stretcher}}} \textcolor{RoyalBlue}{place}. %
\\ \cmidrule{2-3}
&
\textbf{Negative-Arg} &
\textcolor{OrangeRed}{\underline{\textit{An injured man}}} \textbf{transported} \textcolor{OrangeRed}{\underline{\textit{a stretcher}}} \textcolor{RoyalBlue}{in} \textcolor{OrangeRed}{\underline{\textit{protesters}}} \textcolor{RoyalBlue}{instrument}. %
\\

\cmidrule{1-3}
\multirow{8}{*}{\textbf{\makecell[l]{Composed\\ Template}} } & 
\textbf{Template} & 
\textcolor{RoyalBlue}{The image is about} \textbf{\textsc{Transport}}. %
\textcolor{RoyalBlue}{The} \textsc{agent} \textcolor{RoyalBlue}{is} $\langle$arg1$\rangle$. \textcolor{RoyalBlue}{The}  \textsc{entity} \textcolor{RoyalBlue}{is} $\langle$arg2$\rangle$. \textcolor{RoyalBlue}{The} \textsc{instrument}  \textcolor{RoyalBlue}{in} $\langle$arg3$\rangle$. \textcolor{RoyalBlue}{The} \textsc{origin} \textcolor{RoyalBlue}{is} $\langle$arg4$\rangle$. \textcolor{RoyalBlue}{The} \textsc{destination} \textcolor{RoyalBlue}{is} $\langle$arg5$\rangle$.
\\ \cmidrule{2-3}
&
\textbf{Positive} & 
\textcolor{RoyalBlue}{The image is about} \textbf{\textsc{Transport}}. %
\textcolor{RoyalBlue}{The} \textsc{agent} \textcolor{RoyalBlue}{is} {\underline{\textit{protesters}}}. 
\textcolor{RoyalBlue}{The}  \textsc{entity} \textcolor{RoyalBlue}{is} {\underline{\textit{an injured man}}}. 
\textcolor{RoyalBlue}{The} \textsc{instrument} \textcolor{RoyalBlue}{is} {\underline{\textit{a stretcher}}}. %
\\ \cmidrule{2-3}
&
\textbf{Negative-Evt} &
\textcolor{RoyalBlue}{The image is about} \textcolor{OrangeRed}{\textbf{\textsc{Arrest}}}. %
\textcolor{RoyalBlue}{The} \textcolor{OrangeRed}{\textsc{agent}} \textcolor{RoyalBlue}{is} {\underline{\textit{protesters}}}. 
\textcolor{RoyalBlue}{The} \textcolor{OrangeRed}{\textsc{detainee}} \textcolor{RoyalBlue}{is} {\underline{\textit{an injured man}}}. 
\textcolor{RoyalBlue}{The} \textcolor{OrangeRed}{\textsc{place}}  \textcolor{RoyalBlue}{is} {\underline{\textit{a stretcher}}}.  
\\ \cmidrule{2-3}
&
\textbf{Negative-Arg} &
\textcolor{RoyalBlue}{The image is about} \textbf{\textsc{Transport}}. %
\textcolor{RoyalBlue}{The} \textsc{agent} \textcolor{RoyalBlue}{is} \textcolor{OrangeRed}{\underline{\textit{an injured man}}}. 
\textcolor{RoyalBlue}{The}  \textsc{entity} \textcolor{RoyalBlue}{is} \textcolor{OrangeRed}{\underline{\textit{a stretcher}}}. 
\textcolor{RoyalBlue}{The} \textsc{instrument}  \textcolor{RoyalBlue}{is} \textcolor{OrangeRed}{\underline{\textit{protesters}}}. 
\\

\cmidrule{1-3}
\multirow{8}{*}{\textbf{\makecell[l]{Continuous\\ Prompt}} } & 
\textbf{Template} & 
\texttt{\textcolor{RoyalBlue}{[X$_{0}$]}}\textbf{\textsc{Transport}} \textcolor{RoyalBlue}{\texttt{[X$_{1}$]}}\textsc{agent}\textcolor{RoyalBlue}{\texttt{[X$_{2}$]}}$\langle$arg1$\rangle$\textcolor{RoyalBlue}{\texttt{[X$_{3}$]}}\textsc{entity}\textcolor{RoyalBlue}{\texttt{[X$_{2}$]}}$\langle$arg2$\rangle$\textcolor{RoyalBlue}{\texttt{[X$_{3}$]}}\textsc{instrument}\textcolor{RoyalBlue}{\texttt{[X$_{2}$]}}$\langle$arg3$\rangle$\textcolor{RoyalBlue}{\texttt{[X$_{3}$]}}\textsc{origin} \textcolor{RoyalBlue}{\texttt{[X$_{2}$]}}$\langle$arg4$\rangle$ \textcolor{RoyalBlue}{\texttt{[X$_{3}$]}}\textsc{destination} \textcolor{RoyalBlue}{\texttt{[X$_{2}$]}}$\langle$arg5$\rangle$\textcolor{RoyalBlue}{\texttt{[X$_{3}$]}}
\\ \cmidrule{2-3}
&
\textbf{Positive} & 
\textcolor{RoyalBlue}{\texttt{[X$_{0}$]}}\textbf{\textsc{Transport}} \textcolor{RoyalBlue}{\texttt{[X$_{1}$]}}\textsc{agent} \textcolor{RoyalBlue}{\texttt{[X$_{2}$]}}{\underline{\textit{protesters}}} \textcolor{RoyalBlue}{\texttt{[X$_{3}$]}}\textsc{entity} \textcolor{RoyalBlue}{\texttt{[X$_{2}$]}}{\underline{\textit{an injured man}}} \textcolor{RoyalBlue}{\texttt{[X$_{3}$]}}\textsc{instrument} \textcolor{RoyalBlue}{\texttt{[X$_{2}$]}}{\underline{\textit{a stretcher}}} \textcolor{RoyalBlue}{\texttt{[X$_{3}$]}}
\\ \cmidrule{2-3}
&
\textbf{Negative-Evt} &
\textcolor{RoyalBlue}{\texttt{[X$_{0}$]}} 
\textbf{\textcolor{OrangeRed}{\textsc{Arrest}}} \textcolor{RoyalBlue}{\texttt{[X$_{1}$]}} 
\textcolor{OrangeRed}{\textsc{agent}} \textcolor{RoyalBlue}{\texttt{[X$_{2}$]}} 
{\underline{\textit{protesters}}} \textcolor{RoyalBlue}{\texttt{[X$_{3}$]}} \textcolor{OrangeRed}{\textsc{detainee}} \textcolor{RoyalBlue}{\texttt{[X$_{2}$]}} 
{\underline{\textit{an injured man}}} \textcolor{RoyalBlue}{\texttt{[X$_{3}$]}} 
\textcolor{OrangeRed}{\textsc{place}}  \textcolor{RoyalBlue}{\texttt{[X$_{2}$]}}
{\underline{\textit{a stretcher}}} \textcolor{RoyalBlue}{\texttt{[X$_{3}$]}}
\\ \cmidrule{2-3}
&
\textbf{Negative-Arg} &
\textcolor{RoyalBlue}{\texttt{[X$_{0}$]}}\textbf{\textsc{Transport}} \textcolor{RoyalBlue}{\texttt{[X$_{1}$]}}\textsc{agent}  \textcolor{RoyalBlue}{\texttt{[X$_{2}$]}}\textcolor{OrangeRed}{\underline{\textit{an injured man}}}\textcolor{RoyalBlue}{\texttt{[X$_{3}$]}}\textsc{entity}\textcolor{RoyalBlue}{\texttt{[X$_{2}$]}}\textcolor{OrangeRed}{\underline{\textit{a stretcher}}}  \textcolor{RoyalBlue}{\texttt{[X$_{3}$]}}\textsc{instrument} \textcolor{RoyalBlue}{\texttt{[X$_{2}$]}}\textcolor{OrangeRed}{\underline{\textit{protesters}}}\textcolor{RoyalBlue}{\texttt{[X$_{3}$]}}
\\

\cmidrule{1-3}
\multirow{4}{*}{ \textbf{\makecell[l]{Caption\\ Editing}} } & %
\textbf{Positive} & {\underline{\textit{Antigovernment protesters}}} \textbf{carry} {\underline{\textit{an injured man}}} \textcolor{RoyalBlue}{on} {\underline{\textit{a stretcher}}} \textcolor{RoyalBlue}{after clashes with riot police on Independence Square in ...} %
\\ \cmidrule{2-3}
&
\textbf{Negative-Evt} &
{\underline{\textit{Antigovernment protesters}}} \textcolor{OrangeRed}{\textbf{arrest}} {\underline{\textit{an injured man}}} \textcolor{RoyalBlue}{on} {\underline{\textit{a stretcher}}} \textcolor{RoyalBlue}{after clashes with riot police on Independence Square in ...} %
\\ \cmidrule{2-3}
&
\textbf{Negative-Arg} &
\textcolor{OrangeRed}{\underline{\textit{An injured man}}} \textbf{carry} \textcolor{OrangeRed}{\underline{\textit{a stretcher}}} \textcolor{RoyalBlue}{on} \textcolor{OrangeRed}{\underline{\textit{antigovernment protesters}}} \textcolor{RoyalBlue}{after clashes with riot police on Independence Square in ...} %
\\

\cmidrule{1-3}
\multirow{4}{*}{\textbf{\makecell[c]{GPT-3}} } & 
\textbf{Positive} & {\underline{\textit{Protesters}}} \textbf{transported} {\underline{\textit{an injured man}}} \textcolor{RoyalBlue}{with} {\underline{\textit{a stretcher}}}. %
\\ \cmidrule{2-3}
&
\textbf{Negative-Evt} &
{\underline{\textit{Protesters}}} \textcolor{OrangeRed}{\textbf{arrested}} {\underline{\textit{an injured man}}} \textcolor{RoyalBlue}{with} {\underline{\textit{a stretcher}}}. %
\\ \cmidrule{2-3}
&
\textbf{Negative-Arg} &
\textcolor{OrangeRed}{\underline{\textit{An injured man}}} \textbf{transported} \textcolor{OrangeRed}{\underline{\textit{a stretcher}}} \textcolor{RoyalBlue}{and} \textcolor{OrangeRed}{\underline{\textit{protesters}}}. 
\\

\bottomrule
\end{tabular}
\caption{The automatically generated positive and negative descriptions for~\cref{fig:framework}. We use \textbf{bold} to represent events, and \underline{\textit{underline}} stands for arguments. The corrupted event type and arguments are in \textcolor{OrangeRed}{orange}, and templates are  in \textcolor{RoyalBlue}{blue}. \textcolor{RoyalBlue}{\texttt{[X$_i$]}} is learnable prepended token. %
}
\label{table:prompt-example}
\end{table*}

Our goal is to incorporate event structured knowledge into vision-language pretraining. In the following we will address two primary questions regarding the model design: (1) How can the structural event knowledge be acquired? (2) How can the semantics and structures of events be encoded?
We define the symbols used in this paper in \cref{tab:symbols}.

\subsection{Event Structural Knowledge Extraction}
\label{sec:model-ie}
\textbf{Text and Visual Knowledge Extraction.}
We use a state-of-the-art text information extraction system~\cite{lin2020joint, li2020gaia} to extract events of 187 types\footnote{The system uses DARPA AIDA ontology, which is the most fine-grained text event ontology, as attached in the Appendix.
}, covering a wide range of newsworthy events.  
For images, we apply Faster R-CNN~\cite{ren2015faster} trained on Open Images~\cite{kuznetsova2020open} to detect objects. %

\textbf{Primary Event Detection.}
When there are multiple events in the caption, the image typically depicts the primary event of the caption.
We detect the primary event as the event that is closer to the root of dependency parsing tree~\cite{manning2014stanford}, and has a larger number of arguments, higher event type frequency, and  higher similarity between trigger word and the image using the pretrained CLIP model~\cite{radford2021learning}. 
We rank events according to these criteria, and perform majority voting. 
For example, in~\cref{fig:framework}, there are two events \textit{carry} and \textit{clashes} in the caption. We select \textit{carry} as the primary event since it is the root of the dependency tree, %
and it has three arguments, as well as higher similarity with the image. %

\subsection{Event Structure Driven Negative Sampling}
\label{sec:model-semantics}

To force the Text and Vision Encoders to learn robust features about event types and argument roles, %
we design the following strategies to generate challenging negatives. %

\textbf{Negative Event Sampling.} 
We compute the confusion matrix for the event type classifier of the state-of-the-art vision-language pretraining model 
CLIP~\cite{radford2021learning} on the pretraining image-caption dataset. 
The classifier is based on the similarity scores between the event type labels $\phi_v \in \Phi_{V}$ (such as \textsc{Transport}) and the input image $i$, and select the top one as the predicted event type $\phi_{v}^{\star}$. 
\begin{equation*}
    \phi_{v}^{\star} = \argmax_{\phi_v \in \Phi_{V}} \bs{\phi_{v}}^T \cdot \bs{i},
\end{equation*}
where the bold symbols stand for the representations from the Text and Vision Encoders in~\cref{fig:framework}, and we follow CLIP to use Text and Vision Transformers. %
The confusion matrix is computed by comparing the predicted event type with the type of the primary event for the image. 
As a result, the negative event types are the challenging cases in image event typing, i.e., the event types whose visual features are ambiguous with the primary event type. 
For example, in~\cref{fig:framework}, \textsc{Arrest} is sampled as a negative event type, since its visual features are similar to \textsc{Transport}.

\textbf{Negative Argument Sampling.} 
For argument roles, since each event by definition has multiple arguments, we manipulate the order of arguments by 
performing a right-rotation of the argument role sequence. %
In detail, we first order existing argument roles following the ontology definition, %
such as ``\textsc{agent}, \textsc{entity}, \textsc{instrument}'' in \cref{fig:framework}. After that, we right rotate the argument role sequence by one step, resulting in ``\textsc{instrument}, \textsc{agent}, \textsc{entity}''. As a result, each argument is re-assigned to a manipulated role, e.g., \textit{injured man}, the second argument, is manipulated from \textsc{entity} to \textsc{agent}. 
If there is only one argument for the event, we sample a negative role according to the argument confusion matrix %
of the text argument extraction system~\cite{lin2020joint}.%

\textbf{Description Generation.}
To encode the positive and negative event structures using the Text Encoder, we design multiple prompt functions, as shown in~\cref{table:prompt-example}: %
(1) \textbf{Single Template-based Prompt} encodes all arguments in one sentence. %
(2) \textbf{Composed Template-based Prompt} uses a short sentence to each argument. 
(3) \textbf{Continuous Prompt} employs learnable
prepended tokens \texttt{[X$_{i}$]}.
(4) \textbf{Caption Editing} has minimum information loss by only altering event trigger word or switching arguments.
(5) \textbf{GPT-3 based Prompt} generates a semantically coherent natural language description conditioned on the event structure. We employ GPT-3~\cite{floridi2020gpt} and use five manual event description examples as few-shot prompts~\cite{floridi2020gpt} to control the generation. %
The input to GPT-3 is the concatenation of the example events (\texttt{[ex\_v]}) with arguments (\texttt{[ex\_a]}), the example descriptions (\texttt{[ex\_desp]}), and the target events (\texttt{[input\_v]}) with arguments (\texttt{[input\_a]}).
The output of GPT-3 is the target description (\texttt{[output\_desp]}). 
The description is more natural compared to template-based methods. 
\begin{figure}[!h]
  \centering
   \includegraphics[width=1\linewidth]{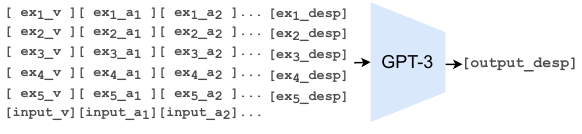}
   \caption{Architecture of GPT-3 based prompt. %
   }
   \label{fig:gpt3}
\end{figure}
\vspace{-5pt}

\begin{table}[t]
\footnotesize
    \centering
    \begin{tabular}{c|m{20em}}
    \toprule 
       Symbol  & Meaning \\
       \midrule
$\langle i, t \rangle$ & image $i$ and its caption text $t$ \\
$o, \phi_o, i_o$         & object, object type, object bounding box \\
$e, \phi_e, t_e$         & entity, entity type, entity text mention \\
$v, \phi_v, t_v$         & event, event type, event text mention  \\
$a \in \mathcal{A}(v)$ & argument role; $\mathcal{A}(v)$ is the Argument role set of event $v$, defined by the IE ontology\textsuperscript{3} \\
$G_{i}$, $G_{t}$ & event graph from image $i$ and text $t$ \\ %
$t^{+}, t^{-}_{v}, t^{-}_{a}$ & positive description, negative event description, negative argument description \\
    \bottomrule
    \end{tabular}
    \caption{List of symbols.
    }
    \vspace{-.3cm}
    \label{tab:symbols}
\end{table}

\subsection{Event Graph Alignment via Optimal Transport}
\label{sec:model-structure}

Each event and its arguments can be organized as a graph, as shown in~\cref{fig:framework}, where the central node is the event node (triangle nodes), and it's connected to entities (circle nodes) via argument roles. 
Encoding event graph structures enables the model to capture the interactions between events and arguments. For example, the \textit{injured man} should be aligned with the \textsc{Entity} being transported, rather than the \textsc{Agent}. %

\textbf{1. Image-level Alignment.}
We compute cosine similarity $s(t, i)$ and distance $d(t, i)$ between the text $t$ and image $i$:
\begin{equation*}
    s(t, i) = \cos(\bs{t} , \bs{i}), 
    d(t, i) = c(\bs{t}, \bs{i}),
\end{equation*}
where $c(\cdot,\cdot)=1-\cos(\cdot,\cdot)$ is the cosine distance function, and $\bs{t}$ is obtained from the Text Transformer and $\bs{i}$ is obtained from the Vision Transformer. 

\textbf{2. Entity-level Alignment.}
The cosine distance between text entity $e$ and image object $o$ considers both the mention similarity and type similarity. 
\begin{equation*}
    d(e, o) = c(\bs{t_{e}}, \bs{i_{o}}) + c(\bs{\phi_{e}}, \bs{\phi_{o}}),
\end{equation*}
where $t_{e}$ is the text mention of entity $e$, and $\bs{t_{e}}$ is its embedding contextualized on the sentence. 
We encode the sentence using the Text Transformer following~\cite{radford2021learning}, and apply average pooling over the tokens in the entity mention $t_{e}$. %
Similarly, $i_{o}$ is the bounding box of object $o$ and $\bs{i_{o}}$ is its embedding contextualized on the image, based on the average pooling over the Vision Transformer representations of the patches covered in the bounding box.
$\phi_{e}$ and $\phi_{o}$ are the type representations encoded by the Text Transformer.
For example, $\phi_{e}=\textsc{person}$ for $e=$ \textit{injured man} and $\phi_{o}=\textsc{person}$ for $o=$ \includegraphics[height=\fontcharht\font`\B]{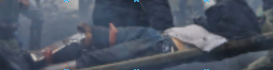}. 
Therefore, the distance between the aforementioned entity and object is:
\begin{equation*}
    d(e, o) = c( \bs{\textit{injured man} ,
    \includegraphics[height=\fontcharht\font`\B]{img/bbox_injured_man.png}} ) 
    + c( \bs{\textsc{Person}} , \bs{\textsc{Person}} ) ,
\end{equation*}

\textbf{3. Event-level Alignment.}
To obtain a global alignment score based on the structures of two graphs, we use the optimal transport~\cite{sinkhorn1964relationship} to get the minimal distance $d(G_{t},G_{i})$ between text event graph $G_{t}$ and image event graph $G_{i}$, %
\begin{equation*}
    d(G_{t},G_{i}) = \min\nolimits_{\bs{T} } \bs{T} \odot \bs{C},  
\end{equation*}
where $\odot$ represents the Hadamard product. $\bs{T} \in \real^{n \times m}_{+}$ denotes the transport plan, learned to optimize a \textit{soft} node alignment between two graphs. $n$ and $m$ are the numbers of nodes in $G_{t}$ and $G_{i}$, respectively.  Namely, each node in text graph $G_{t}$ can be transferred to multiple nodes in image graph $G_{i}$ with different weights. 

$C$ is the cost matrix. We define cost between event nodes, and between argument nodes. 
For event nodes, the cost is the cosine distance between the image $i$ and trigger word $v$, 
\begin{equation*}
    C(v, i) = c(\bs{t_{v}} , \bs{i}) + c(\bs{\phi_{v}} , \bs{i} ).
\end{equation*}
For example, in~\cref{fig:framework}, $v$\ $=$\ \textit{carry} and $\phi_{v}$\ $=$\ \textsc{Transport}, 
\begin{equation*}
    C(v, i) = c(\textit{carry} , \includegraphics[height=\fontcharht\font`\B]{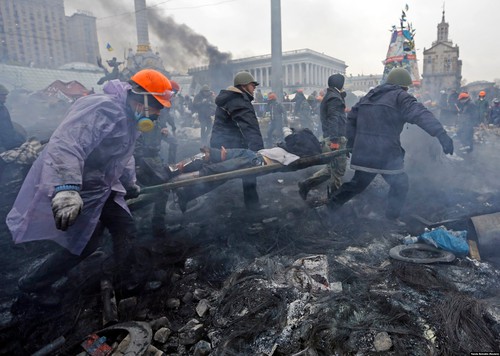}) + c(\textsc{Transport} , \includegraphics[height=\fontcharht\font`\B]{img/carry.jpeg} ).
\end{equation*}
The representation $\bs{t}_v$ is also from the Text Transformer, contextualized on the text sentence. 

The cost between each argument $\langle a, e\rangle$ and each bounding box $o$ is based on the similarity of object $o$ with both argument role $a$ and text entity $e$. 
\begin{align*}
    C(\langle a, e\rangle, o) & 
    = d(a, o) + d(e, o) 
    \\
    & = c( \bs{t_{a}} , \bs{i_{o}} ) 
    + c( \bs{t_{e}} , \bs{i_{o}} ) 
    + c( \bs{\phi_{e}} , \bs{\phi_{o}} )
    ,
\end{align*}
where ${t_{a}}$ is the argument description. %
For example, for the argument role $a=\textsc{Entity}$ of entity $e=$ \textit{injured man},
\begin{align*}
    C( &  \langle a, e\rangle, o)   
      = c( \textit{\textsc{Entity} of \textsc{Transport}} ,%
      \includegraphics[height=\fontcharht\font`\B]{img/bbox_injured_man.png} ) 
    \\
    & + c( \bs{\textit{injured man}} ,
    \includegraphics[height=\fontcharht\font`\B]{img/bbox_injured_man.png} ) 
    + c( \bs{\textsc{Person}} , \bs{\textsc{Person}} ) .
\end{align*}

The optimal $\bs{T} \in \real_{+}^{n \times m}$ that solves $d(G_{t},G_{i}) = \min\nolimits_{\bs{T} } \bs{T} \odot \bs{C}$ 
can be approximated by a differentiable Sinkhorn-Knopp algorithm~\cite{sinkhorn1964relationship,cuturi2013sinkhorn} following~\cite{xu2019scalable}, 
\begin{equation*}
    \bs{T} = \text{diag}(\bs{p}) \ \text{exp}(-\bs{C}/\gamma) \ \text{diag}(\bs{q}),
\end{equation*}
where $\bs{p} \in \real_{+}^{n \times 1}$ and $\bs{q} \in \real_{+}^{m \times 1}$. 
Starting with any positive vector $\bs{q}^{0}$ to perform the following iteration: 
\begin{equation*}
\begin{aligned}
    \text{for } \ & i = 0, 1,2,\dots \text{until convergence}, \\
     & \bs{p}^{i+1} = \mathbf{1} \oslash (\bs{K}\bs{q}^{i}),\ \  %
    \bs{q}^{i+1} = \mathbf{1} \oslash (\bs{K}^{\top} \bs{p}^{i+1}),
\end{aligned}
\end{equation*}
where $\oslash$ denotes element-wise division. $\bs{K} = \text{exp}(-\bs{C}/\gamma) $. 
A computational $\bs{T}^{k}$ can be obtained by iterating for a finite number $k$ times, 
\begin{equation*}
   \bs{T}^{k} :=  \text{diag}(\bs{p}^{k})\bs{K}\text{diag}(\bs{q}^{k}).
\end{equation*}
\subsection{Contrastive Learning Objective}
\label{sec:model-constrastive}

We optimize the cosine similarity between image $i$ and positive description $t^{+}$ to be close to $1$, while negative descriptions $t^{-}$ to be close to $0$,
\begin{equation*}
   L_{1}=\sum\nolimits_{\langle t,i \rangle} D_{KL} ( s(t,i) \ || \ \mathds{1}_{t \in T^{+}} ),
\end{equation*}
where $D_{KL}(\cdot || \cdot)$ is the Kullback-Leibler divergence, and $\mathds{1}_{t \in T^{+}}$ is the indicator function showing whether the description is a positive description.
It enables our model to handle any number of positive and negative descriptions.
Also, we include the descriptions of other images in the same batch as negative descriptions.

We also minimize the distance between two event graphs,
\begin{equation*}
   L_{2}=\sum\nolimits_{\langle t,i \rangle} d(G_{t},G_{i}).
\end{equation*}
The contrastive learning of event and argument description and the alignment of event graphs are jointly optimized:
\begin{equation*}
   L=\lambda_{1} L_{1} + \lambda_{2} L_{2}.
\end{equation*}
We set $\lambda_{1}$ and $\lambda_{2}$ as $1$ in this paper.

\begin{figure*}[ht]

  \centering
  \begin{subfigure}{0.49\linewidth}
    \includegraphics[width=1.0\linewidth]{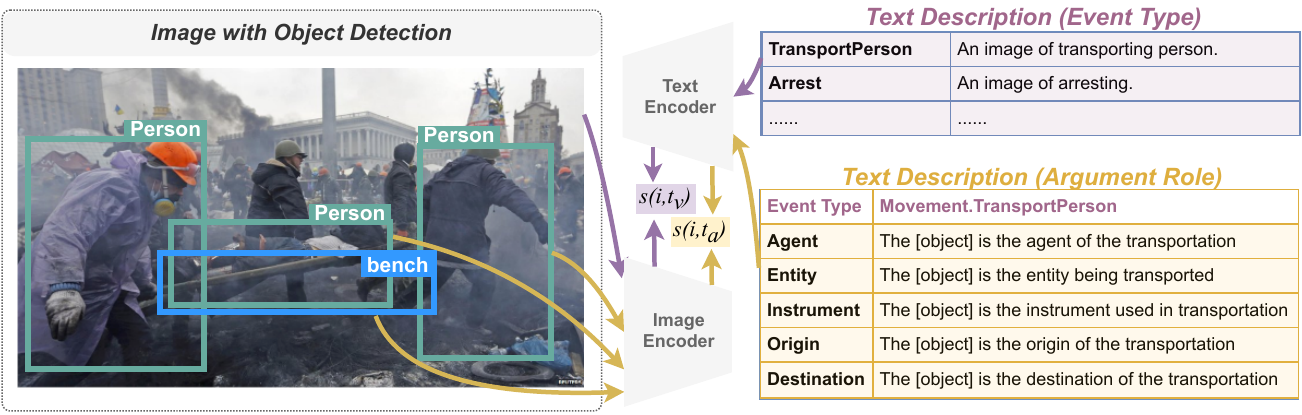}
    \caption{Architecture of event extraction (\mmee and GSR). Event typing (in \textcolor{Violet}{purple}) ranks event type descriptions given the image, and argument extraction (in \textcolor{YellowOrange}{yellow}) rank argument descriptions given the bounding box. %
    }
    \label{fig:evalution-m2e2}
    \vspace{-5pt}
  \end{subfigure}
  \hfill
  \begin{subfigure}{0.49\linewidth}
    \includegraphics[width=1.0\linewidth]{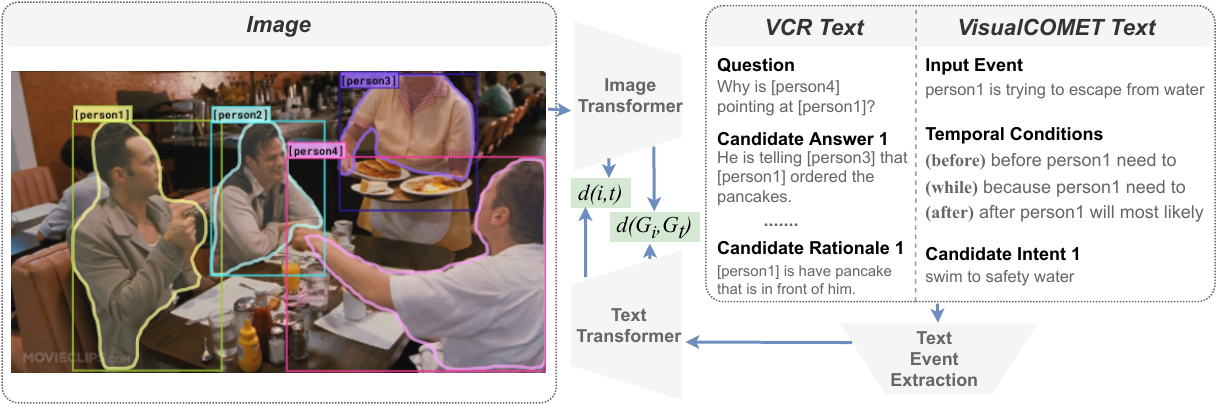}
    \caption{Architecture of VCR and VisualCOMET. We rank $\langle$question, answer$\rangle$ and $\langle$input event, temporal condition, intent$\rangle$ respectively given the image. %
    We calculate both the image-text level alignment and the event graph alignment.
    }
    \label{fig:evalution-vcr}
    \vspace{-5pt}
  \end{subfigure}
   \caption{Architecture for evaluation tasks. %
   }
   \label{fig:evalution}
   \vspace{-.2cm}
\end{figure*}

\section{Evaluation Tasks}
\label{sec:evaluation}

\subsection{Multimedia Event Extraction (\mmee)} 
\label{sec:task_m2e2}

\noindent
\textbf{Task Setting.} 
Multimedia Event Extraction~\cite{li2020cross} aims to (1) classify images into eight event types, and (2) localize argument roles as bounding boxes in images. We choose this task as a direct assessment of event structure understanding.

\noindent
\textbf{Our Approach.} 
\underline{Zero-shot Setting:}
We evaluate the models' ability to handle open vocabulary events, as required by real-world applications. Also, zero-shot evaluation provides a direct comparison of the effectiveness of event knowledge encoding during pretraining. 
As shown in \cref{fig:evalution-m2e2}, we select the event type with the highest similarity score $s(i,t)$ with the image, and for each bounding box, we rank candidate argument roles of the selected event type. %
\underline{Supervised Setting:}  %
We include the supervised setting to prove the effectiveness of the model architecture at encoding event knowledge in the presence of direct supervision, with details in Appendix A.3.

\noindent
\textbf{Evaluation Metrics.}
We follow~\cite{li2020cross} to use F-scores to evaluate event typing and argument extraction. %

\subsection{Grounded Situation Recognition (GSR)} 
\label{sec:task_gsr}

\noindent
\textbf{Task Setting.}
Grounded Situation Recognition~\cite{pratt2020grounded} selects an event type from 504 verbs, and predicts the entity name and the bounding box for each argument role. %

\noindent
\textbf{Our Approach.} 
The implementation is similar to \mmee in~\cref{fig:evalution-m2e2}, with details in Appendix A.4.

\noindent
\textbf{Evaluation Metrics.} We follow~\cite{pratt2020grounded} as detailed in the Appendix. %

\subsection{Image Retrieval} 
\label{sec:task_image_retrieval}

\noindent
\textbf{Task Setting.}
Image retrieval ranks images for each given caption, which is a direct evaluation on the image-text alignment. %

\noindent
\textbf{Our Approach.}
We perform the alignment of image and text $d(i,t)$, and event graphs across two modalities $d(G_{i}, G_{t})$.

\noindent
\textbf{Evaluation Metrics.}
We use conventional image retrieval measures including \textit{Recall@1}, \textit{Recall@5} and \textit{Recall@10}.

\subsection{Visual Commonsense Reasoning (VCR)} 
\label{sec:task_vcr}

\noindent
\textbf{Task Setting.}
Given a question, the task contains two subtasks: (1) \textit{Answer Prediction} from four options; (2) \textit{Rationale Prediction} from four options to support the answer.

\noindent
\textbf{Our Approach.}
To evaluate the quality of pretraining models, as shown in \cref{fig:evalution-vcr}, we adopt zero-shot settings solely relying on image-text alignment for a fair comparison, with details described in Appendix A.5.

\noindent
\textbf{Evaluation Metrics.}
We use F-scores to evaluate both of answer prediction and rationale prediction, following~\cite{zellers2019recognition}.

\subsection{Visual Commonsense Reasoning in Time} 
\label{sec:visualcomet}

\noindent
\textbf{Task Setting.}
Given an image and the event involved with its participants, VisualCOMET~\cite{park2020visualcomet} aims to generate ``{intent}s'' of the participants, as detailed in Appendix A.6. %

\noindent
\textbf{Our Approach.}
As shown in \cref{fig:evalution-vcr}, we rank candidate intents based on the image-text similarity (Appendix A.6). %

\noindent
\textbf{Evaluation Metrics.}
We adopt \textit{Accuracy@50} following the perplexity evaluation of the state-of-the-art model~\cite{park2020visualcomet}.

\section{Experiments}

\subsection{Pretraining Details}

\noindent
\textbf{A New Dataset.}
\label{sec:dataset}
We collect 106,875 image-captions that are rich in events from news websites~\cite{voa}.
It provides a new challenging image-retrieval benchmark, where each sentence may contain multiple events with a complicated linguistic structure. The average caption length is 28.3 tokens, compared to 13.4 for Flickr30k and 11.3 for MSCOCO.
The data statistics are shown in~\cref{table:data_stats}, with structural event knowledge extracted automatically following~\cref{sec:model-ie}.

\begin{table}[!hbt]
\footnotesize
\centering
\setlength\tabcolsep{5pt}
\begin{tabular}{c c ccccc}
\toprule
\textbf{Dataset} & \textbf{Split} & \textbf{\#image}  & \textbf{\#event} & \textbf{\#arg} & \textbf{\#ent} \\ %
\midrule
 &\textbf{Train} &   76,256 & 84,120 & 148,262 & 573,016  \\
 \textbf{VOANews} 
 &\textbf{Test} & 18,310 & 21,211 & 39,375 & 87,671\\
 & \textbf{No-event} & 12,309 & - & - & -  \\
\bottomrule
\end{tabular}
\caption{Data statistics of VOANews.  
}
\label{table:data_stats}
\end{table}

\begin{table*}[!hbt]
\footnotesize
\centering
\setlength\tabcolsep{5.5pt}
\setlength\extrarowheight{2pt}
\begin{tabular}{l | l | p{2em}p{2em}p{2em}|p{2em}p{2em}p{2em} | c|cccc}  %
\toprule
& %
&
\multicolumn{6}{c|}{\textbf{Multimedia Event Extraction (M\textsuperscript{2}E\textsuperscript{2})}} &
\multicolumn{5}{c}{\textbf{Grounded Situation Recognition (SWiG)}} 
\\
\multirow{1}{*}{\textbf{Setting}}
& \textbf{Model} & 
\multicolumn{3}{c|}{\textbf{Event}} & \multicolumn{3}{c|}{\textbf{Argument}} &
\multicolumn{1}{c|}{\textbf{Event}} & \multicolumn{4}{c}{\textbf{Argument}} 
\\ 
& 
& { P } & { R } & { F\textsubscript{1} } 
& { P } & { R } & { F\textsubscript{1} } 
& {verb} & {value} & {value-all} & {ground} & {ground-all}
\\
\midrule
\multirow{4}{*}{ {Zero-shot}}

& {CLIP} 
& 29.5 & 65.7 & 40.7 & 9.2 & 12.7 & 10.7
& 28.3 & 13.3 & 7.6 & 11.2 & 3.8
\\ 

& {CLIP pretrained on news}
& 31.7 & 64.7 & 42.6 & 9.7 & 13.1 & 11.1
& 29.9 & 14.0 & 8.2 & 12.0 & 4.3
\\\cmidrule{2-13}

& \textbf{CLIP-Event}
& 36.4 & 70.8 & \textbf{48.1} & {13.9} & {16.0} & \textbf{14.8}
& \textbf{31.4} & \textbf{14.9} & \textbf{9.2} & \textbf{12.8} & \textbf{5.2}
\\
& \ \ \  w/o OptimalTransport
& 35.0 & 59.3 & 44.1 & 11.0 & 12.6 & 11.9  
& 30.2 & 14.2 & 8.4 & 12.3 & 4.4
\\ 
& \ \ \  Single Template
& 32.3 & 71.4 & 44.4 & 11.9 & 15.6 & 13.2
& 30.4 & 14.4 & 8.6 & 12.4 & 4.7
\\
& \ \ \  Composed Template
& 33.9 & 72.8 & 46.3 & 12.7 & 15.3 & 13.9
& 30.9 & 14.5 & 8.8 & 12.4 & 4.8
\\
& \ \ \  Continuous Prompt
& 33.6 & 75.7 & 46.5 & 11.1 & 16.7 & 13.3
& 30.4 & 14.0 & 8.3 & 12.1 & 4.3
\\
& \ \ \  Caption Editing
& 30.9 & 71.4 & 43.2 & 11.6 & 13.8 & 12.6
& 30.1 & 13.9 & 8.2 & 12.3 & 4.4
\\
& \ \ \  GPT-3 Prompt
& 34.2 & 76.5 & 47.3 & 12.1 & 16.8 & 14.1
& 31.1 & 14.9 & 9.1 & 12.7 & 5.2
\\
\midrule
\multirow{4}{*}{ {Supervised}}
& {State-of-the-Art~\cite{li2020cross, pratt2020grounded}} 
& 43.1 & 59.2 & 49.9 & 14.5 & 10.1 & 11.9
& 39.9 & 31.4 & 18.9 & 24.9 & 9.7
\\
& {CLIP finetuned on SWiG}
& 38.1 & 71.6 & 49.8 & 20.9 & 12.8 & 15.9
& 42.6 & 32.6 & 19.2 & 25.2 & 10.2
\\\cmidrule{2-13}
& \textbf{CLIP-Event\textsuperscript{+SWiG}}
& {41.3} & {72.8} & \textbf{52.7} & {21.1} & {13.1} & \textbf{17.1}
& \textbf{45.6} & \textbf{33.1} & \textbf{20.1} & \textbf{26.1} & \textbf{10.6}
\\
& \ \ \  w/o OptimalTransport
& 40.3 & 71.3 & 51.5 & 20.8 & 13.0 & 16.0
& 44.7 & 32.9 & 19.4 & 24.4 & 10.1

\\
\bottomrule
\end{tabular}
\caption{Evaluation results and ablation studies on image event extraction. We follow the evaluation measures (\%) of each benchmark. 
}
\vspace{-.1cm}

\label{table:result_extraction}
\end{table*}

\begin{table}[!hbt]
\footnotesize
\centering
\setlength\tabcolsep{4.5pt}
\setlength\extrarowheight{2pt}
\begin{tabular}{ l  | c c| c c| c c} %
\toprule
\multirow{1}{*}{\textbf{Model}} & \multicolumn{2}{c|}{\textbf{Flickr30k}} & \multicolumn{2}{c|}{\textbf{MSCOCO}} & \multicolumn{2}{c}{\textbf{VOANews}}
\\
\midrule
{CLIP}
& 62.2 & 81.9 & 30.3 & 50.3 & 21.2 & 23.4
\\
{CLIP} pretrained on news
& 64.3 & 81.2 & 32.2 & 50.8 & 23.5 & 25.1
\\
\midrule
\textbf{CLIP-Event}
& \textbf{67.0} & \textbf{82.6} & \textbf{34.0} & \textbf{51.3} & \textbf{27.5} & \textbf{28.7}
\\
\ \ \  w/o OptimalTransport
& 65.6 & 80.5 & 32.5 & 51.0 & 25.5 & 26.9
\\
\bottomrule
\end{tabular}
\vspace{-.1cm}
\caption{
R@1(\%) on text-to-image (left) and image-to-text (right) retrieval on Flickr30k (1k test), MSCOCO (5k test) and VOANews. %
}
\label{table:result_retrieval}
\end{table}

\begin{table}[!hbt]
\vspace{-.4cm}
\footnotesize
\centering
\setlength\tabcolsep{3.5pt}
\setlength\extrarowheight{2pt}
\begin{tabular}{ l  | c c| c} %
\toprule
\multirow{2}{*}{\textbf{Model}} & \multicolumn{2}{c|}{\textbf{VCR}} & \textbf{VisualCOMET} 
\\
 & {Answer F\textsubscript{1}} & {Rationale F\textsubscript{1}} & {Accuracy@50} 
\\
\midrule
{Perplexity in \cite{park2020visualcomet} }
& - & - & 18.2
\\
{CLIP}
& 51.1 & 46.8 &  20.1
\\
{CLIP} pretrained on news
& 51.8 & 47.2 & 20.9
\\
\midrule
\textbf{CLIP-Event}
& \textbf{52.4} & \textbf{49.2} & \textbf{22.4}
\\
\ \ \  w/o OptimalTransport
& 52.0 & 48.6 & 21.1
\\
\bottomrule
\end{tabular}
\caption{Results (\%) on zero-shot VCR and VisualCOMET. %
}
\vspace{-.3cm}
\label{table:result_vcr}
\end{table}

\noindent
\textbf{Parameter Settings.}
We utilize the Text and Vision Transformers of ``ViT-B/32'' to initialize our encoders. %
More details are included in the Appendix.

\subsection{Baselines}
\noindent
\textbf{State-of-the-art Multimedia Pretraining Models.}
We compare with CLIP~\cite{radford2021learning} by
running the public release of ``ViT-B/32" and report the scores in the following experiments for a fair comparison. %
We further pretrain CLIP using the image-captions in the same dataset in \cref{table:data_stats} for a fair comparison in terms of data resources.

\noindent
\textbf{State-of-the-art Event Extraction Models.}
The state-of-the-art event extraction models, such as WASE~\cite{li2020cross} for Multimedia Event Extraction task, JSL~\cite{pratt2020grounded} for Grounded Situation Recognition task. 

\noindent
\textbf{Ablation Study: CLIP-Event w/o Optimal Transport} is included as a variant of our model in which we remove the alignment between event graphs. It is trained only on the contrastive loss $L_{1}$.

\noindent
\textbf{Ablation Study: Each Prompt Function} is used solely during training, for the purpose of comparing its effectiveness.

\subsection{Analysis on Event Extraction Tasks}

Under zero-shot settings, we achieve %
5.5\% absolute F-score gain on event extraction, and 33.3\% relative gain on argument extraction on \mmee, as shown in \cref{table:result_extraction}. 

The gains achieved by pretraining on news data are significantly amplified with the help of structural event knowledge. For example, CLIP pretrained on news achieves 1.9\% improvement compared to the vanilla CLIP on \mmee. Our CLIP-Event significantly boosts the gain to 3.89 times.

Zero-shot CLIP-Event outperforms the state-of-the-art weakly supervised model on argument extraction on \mmee dataset, showing that the proposed optimal transport alignment effectively captures the argument structures, which previous vision-language pretraining models fail. %

\begin{figure}[!h]

  \centering
  \begin{subfigure}{0.49\linewidth}
    \includegraphics[width=1.0\linewidth, height=12em]{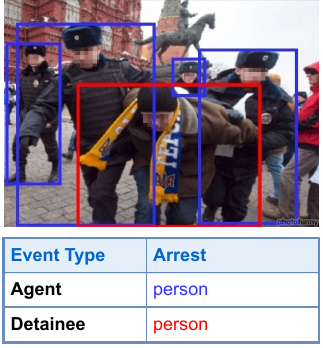}
    \caption{An example result on \mmee. 
    }
    \label{fig:result-m2e2}
    \vspace{-8pt}
  \end{subfigure}
  \hfill
  \begin{subfigure}{0.47\linewidth}
    \includegraphics[width=1.0\linewidth,height=12em]{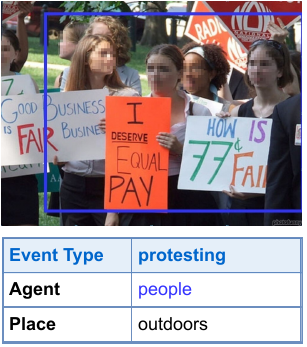}
    \caption{An example result on SWiG.
    }
    \label{fig:result-swig}
    \vspace{-8pt}
  \end{subfigure}
   \caption{Example results of event extraction tasks.
   }
   \label{fig:result_event}
\end{figure}

For argument localization, CLIP-Event achieves a higher gain on \mmee than SWiG, due to the fact that SWiG uses a different argument bounding box grounding strategy. SWiG merges all objects that play the same role into a single large bounding box. As shown in \cref{fig:result-swig}, our approach detects argument roles for each object first, and then merges those objects of the same role into the a large bounding box. In comparison, \mmee allows multiple objects with the same argument role, which is consistent with our approach to use objects aligning with argument roles, as shown in \cref{fig:result-m2e2}.

\subsection{Analysis on Downstream Tasks}

\begin{figure}[!h]
  \centering
    \includegraphics[width=1.0\linewidth]{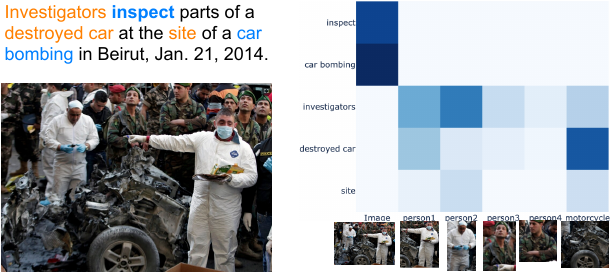}
  \caption{Example results of text-to-image retrieval on VOANews, with the visualizations of the optimal transport plan.}
  \label{fig:result-voa}
\end{figure}

\noindent
\textbf{Image Retrieval.} 
(1) VOANews presents a greater challenge due to the various events in the captions and the more difficult sentence structures compared to Flickr30k and MSCOCO, as shown in \cref{fig:result-voa}. 
The improvement on VOANews is much higher than the gains on Flickr30k and MSCOCO, proving that our model is capable of handling lengthy sentences, particularly those with many events.

\noindent
(2) Downstream tasks benefit from fine-grained event graph alignments. For example, in \cref{fig:result-voa}, the strong alignment between objects and \textit{investigators} and \textit{destroyed car} enables the image to be successfully ranked higher.

\begin{figure}[!h]
  \centering
    \includegraphics[width=1.0\linewidth]{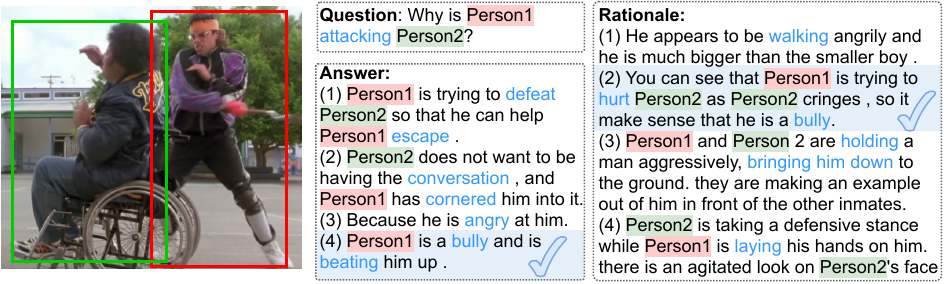}
   \caption{VCR can benefit from event (in \textcolor{RoyalBlue}{blue}) understanding.
   }
   \label{fig:result_vcr}
\end{figure}

\noindent
\textbf{VCR.}
(1) On VCR, the rationale F\textsubscript{1} improves more than answer F\textsubscript{1}. Rationale prediction is more challenging since it refers to the details of the scene, %
which our fine-grained alignment well captures.
(2)
Event knowledge is particularly beneficial for downstream tasks.
In \cref{fig:result_vcr}, only the correct answer corresponds to the event type of the input image.

\noindent
\textbf{VisualCOMET.}
We compare our results to the perplexity of the state-of-the-art model, %
which is also retrieval-based. The baseline is trained using the training set of VisualCOMET, but our model is an unsupervised model, which achieves superior performance, demonstrating that our model is capable of comprehending events in the images.

\subsection{Ablation Studies}

\noindent
\textbf{Effect of Event Graph Alignment via Optimal Transport.}
(1) Removing optimal transport (``w/o OptimalTransport'') generally lowers the performance on all evaluation tasks, since it ignores the event graph structures and their cross-media alignment, but relies solely on the overly simplistic image and sentence features. 
(2) The performance gain on argument extraction task is the highest, since it requires the fine-grained alignment of text and images. 
(3) We visualize the transport plan in \cref{fig:result-voa} to bring insights into the learned alignment. 
It is a global decision that takes the argument structures of two event graphs into account. Thus, distinct argument roles tend to be associated with diverse objects with different visual features in order to achieve a low \textit{global} transport cost. 
For instance, \textit{investigators} match objects dressed in white, but not soldier objects, due to the dissimilar visual features. 
Additionally, one argument role tends to be aligned with objects that have similar visual features, e.g., two \textit{investigators} are both dressed in white protection suits.

\noindent
\textbf{Comparison between prompt functions.}
As shown in \cref{table:result_extraction}, GPT3 provides the optimal performance among prompt functions. It leverages the knowledge encoded in GPT3, thus generating natural descriptions with precise event information.  %
Other prompt functions also demonstrate their effectiveness in supporting event understanding.

\section{Related Work}

\textbf{Vision-Language Pretraining.}
Recent years have witnessed great success in Vision-Language pretraining models~\cite{lu2019vilbert,tan2019lxmert,chen2020uniter,li2020oscar,zhou2020unified,zhang2021vinvl, radford2021learning,jia2021scaling,kim2021vilt,huang2021seeing,wang2021simvlm} based on Transformer architectures~\cite{vaswani2017attention}. %
Image structures have been proven useful to pretraining models, %
such as scene graphs~\cite{yu2020ernie}. However, event structural knowledge is not well captured in pretraning models, %
demonstrating deficiencies in tasks related to verb comprehension%
~\cite{hendricks2021probing}. We are the first to encode structural event knowledge to enhance vision-language pretraining. %

\textbf{Visual Event Understanding.}
Previous work simplifies visual events as \textit{verbs}  using Subject-Verb-Object triples ~\cite{yao2010modeling,sung2012unstructured,das2013thousand,chao2015hico,gupta2015visual,lu2016visual,shekhar2017foil,kato2018compositional,li2019transferable,wang2020learning,zhou2020cascaded}. 
Situation Recognition~\cite{yatskar2016situation, pratt2020grounded} aims to detect argument roles and 
Multimedia Event Extraction~\cite{li2020cross} categorizes verbs into event types. However, their limited event ontologies fail to handle open-world events in real applications. %
In contrast, our proposed pretraining model supports zero-shot event extraction and demonstrate good performance on other downstream tasks requiring image event reasoning. 

\textbf{Cross-media Alignment.}
Existing pretraining models~\cite{tan2019lxmert,chen2020uniter,li2020oscar, zhang2021vinvl,chen2020graph} maximize the alignment across two modalities without taking into account the structure of text and images. %
Image structures~\cite{zareian2020weakly,li2020cross} that are analogous to text linguistic structures are proposed. %
There is, however, a gap between complicated linguistic structures and image structures. We propose to use the  text event graph structures to fill in the gap and compute a global alignment over two event graphs.

\section{Conclusions and Future Work}
This paper proposes to integrate structural event knowledge into vision-language pretraining. 
We perform cross-media transfer of event knowledge, by automatically extracting event knowledge from captions and supervising image event structure understanding via contrastive learning. 
We generate hard negatives by manipulating event structures based on confusion matrices, and design event prompt functions to encode events into natural sentences. 
To transfer argument structural knowledge, we propose an event graph alignment loss via optimal transport, obtaining a global alignment based on argument structures. 
It outperforms the state-of-the-art vision-language pretraining models on event extraction and downstream tasks under zero-shot settings. 
In the future, we will expand this capability to videos to comprehend the evolution of events using argument tracking.

\section*{Acknowledgement}
We thank the anonymous reviewers helpful suggestions. 
This research is based upon work supported by U.S. DARPA AIDA Program No. FA8750-18-2-0014, U.S. DARPA KAIROS Program No. FA8750-19-2-1004. The views and conclusions contained herein are those of the authors and should not be interpreted as necessarily representing the official policies, either expressed or implied, of DARPA, or the U.S. Government. The U.S. Government is authorized to reproduce and distribute reprints for governmental purposes notwithstanding any copyright annotation therein.

\appendix

\section{Implementation Details}

\subsection{Event Knowledge Extraction Details.}

\textbf{Text Knowledge Extraction Details.}
We use the state-of-the-art text information extraction tools OneIE~\cite{lin2020joint}. In detail, we run the dockerized version GAIA~\cite{ li2020gaia} that is using the DARPA AIDA event ontology~\footnote{ \url{https://github.com/NextCenturyCorporation/AIDA-Interchange-Format/blob/master/java/src/main/resources/com/ncc/aif/ontologies/LDCOntologyM36}}, the most fine-grained text event ontology, attached in \textit{event\_ontology\_oneie.json}. 
\begin{table}[!ht]
\tiny
    \centering
    \begin{tabular}{l|l}
    \toprule 
       Example Event Types & Arguments \\
       \midrule
ArtifactExistence.ArtifactFailure.MechanicalFailure 
& MechanicalArtifact, Instrument, Place
\\
ArtifactExistence.DamageDestroy.Damage
& Damager, Artifact, Instrument, Place
\\
ArtifactExistence.DamageDestroy.Destroy
& Destroyer, Artifact, Instrument, Place
\\
ArtifactExistence.Shortage.Shortage
& Experiencer, Supply, Place
\\
Conflict.Attack
& Attacker, Target, Instrument, Place
\\
Conflict.Attack.AirstrikeMissileStrike
& Attacker, Target, Instrument, Place
\\
Conflict.Attack.BiologicalChemicalPoisonAttack
& Attacker, Target, Instrument, Place
\\
Conflict.Attack.Bombing
& Attacker, Target, Instrument, Place
\\
Conflict.Attack.FirearmAttack
& Attacker, Target, Instrument, Place
\\
Conflict.Attack.Hanging
& Attacker, Target, Instrument, Place
\\
Conflict.Attack.Invade
& Attacker, Target, Instrument, Place
\\
Conflict.Attack.SelfDirectedBattle
& Attacker, Target, Instrument, Place
\\
Conflict.Attack.SetFire
& Attacker, Target, Instrument, Place
\\
Conflict.Attack.Stabbing
& Attacker, Target, Instrument, Place
\\
Conflict.Attack.StealRobHijack
& Attacker, Target, Instrument, Place
\\
Conflict.Attack.Strangling
& Attacker, Target, Instrument, Place
\\
Disaster.AccidentCrash.AccidentCrash
& DriverPassenger, Vehicle, CrashObject, Place
\\
Disaster.DiseaseOutbreak.DiseaseOutbreak
& Disease, Victim, Place
\\
Disaster.FireExplosion.FireExplosion
& FireExplosionObject, Instrument, Place
\\
Justice.ArrestJailDetain.ArrestJailDetain
& Jailer, Detainee, Crime, Place
\\
Justice.InitiateJudicialProcess
& Prosecutor, Defendant, JudgeCourt, Crime
\\
Justice.InitiateJudicialProcess.ChargeIndict
& Prosecutor, Defendant, JudgeCourt, Crime
\\
Justice.InitiateJudicialProcess.TrialHearing
& Prosecutor, Defendant, JudgeCourt, Crime
\\
Justice.Investigate
& Investigator, Defendant, Place
\\
... & ...
\\
    \bottomrule
    \end{tabular}
    \caption{Example event types from Text Information Extraction system, the full list is attached in \textit{event\_ontology\_oneie.json}.
    }
    \label{tab:ontology}
\end{table}

In addition, we explore open-world event extraction that is not limited to a specific event ontology. We apply OpenIE tools~\cite{schmitz2012open, angeli2015leveraging}, which output $\langle$\textit{subject}, \textit{relation}, \textit{object}$\rangle$. For example, from the caption in Fig. 2 in the main paper, OpenIE extracts $\langle$\textit{protesters}, \textsc{carry}, \textit{injured man}$\rangle$, $\langle$\textit{clashes}, \textsc{with}, \textit{riot police}$\rangle$, and  $\langle$\textit{Independence Square}, \textsc{in}, \textit{Kyiv}$\rangle$. However, from 100 randomly selected captions, we find that 72.1\% events from OpenIE are not visually detectable, such as \textsc{thinking} and \textsc{inviting}. Considering that these events will introduce a lot of noise to the cross-media alignment, we only adopt the aforementioned supervised IE model to obtain event knowledge from text.  

\textbf{Visual Knowledge Extraction Details.}
We apply Faster R-CNN~\cite{ren2015faster} to detect objects, which is trained on Open Images~\cite{kuznetsova2020open} with 600 object types (classes). 
For event knowledge extraction on images, the most similar tool is grounded situation recognition~\cite{pratt2020grounded}%
, which achieves 39.6\% accuracy on event extraction. %
Considering the errors propagated from extraction models, instead of extracting event knowledge from images as a supervision signal, we take advantage of text information extraction that have better event extraction performance (75.2\% on F-score), to provide supervision to enhance  visual event understanding.

\subsection{Parameter Settings}
We utilize the Text and Vision Transformers of ``ViT-B/32'' to initialize our encoders. The batch size is $128$. We set the learning rate as $1e-6$ with a linearly-decaying schedule. We train $20$ epochs with Adam~\cite{loshchilov2018fixing} as the optimizer, and select the best model based on the image-retrieval performance on VOANews testing dataset.  
The optimal transport plan is obtained within $k=50$ iterations. 
To get the bounding box embeddings from CLIP visual backbone, we extract grid features and perform average pooling on the grids covered by the bounding box. For CLIP-ViT-B models, we reshape the patch representation of the final layer into grid features. For CLIP-ResNet models, we use the grid features from the last layer before the pooling.
The model is trained on eight Tesla V100 GPUs with 32GB DRAM, and the pretraining takes around one day.

\subsection{Multimedia Event Extraction Implementation Details}
\label{app:m2e2}

\textbf{Task Setting.} 
Multimedia Event Extraction~\cite{li2020cross} aims to (1) classify images into eight event types, and (2) localize argument roles as bounding boxes in images. 

\textbf{Evaluation Goal.}
We choose this task as a direct assessment of event structure understanding.

\textbf{Our Approach.} 
Under zero-shot settings, we directly evaluate the pretraining model on the testing set. We evaluate the event extraction and argument extraction on all images, which contain visual events of 8 types. We add \textsc{Other} to detect the images not belonging to the eight target types. The description of \textsc{Other} is \textit{An image of other events.} For argument extraction, we rank argument roles for each object bounding box, and also add \textsc{Other} argument role as a candidate with the description \textit{other roles of the event}. 

Under supervised settings, we use the same training data SWiG as the sate-of-the-art model~\cite{li2020cross}, but replacing the text event table with the annotation table, and setting the optimal transport plan as the fine-grained alignment between event graphs. 
We use the same training dataset SWiG~\cite{pratt2020grounded} with 125k images to further finetune our model to compare with the supervised models. During finetuning, we replace the text event extraction results with the annotated events for images, and set the optimal transport plan as the ground truth alignment between arguments and object bounding boxes. 

\subsection{Grounded Situation Recognition Implementation Details}
\label{app:gsr}

\textbf{Task Setting.}
Grounded Situation Recognition~\cite{pratt2020grounded} selects an event type from 504 verbs, and predicts the entity name and the bounding box for each argument role. 

\textbf{Evaluation Goal.}
It is also a direct evaluation of event structure understanding, but with larger size of event types and argument roles.

\textbf{Implementation.}
Grounded Situation Recognition requires the model to assign each image to a verb from 504 verbs (such as \textsc{Riding}), and name the argument (such as \textit{man}) of each argument role (such as \textsc{Agent}). For each image, we rank the verbs using the description ``\textit{An image of} $\langle$\textit{verb}$\rangle$''. For each argument role, we obtain the candidate names from the training set, and rank the candidate names using the description ``\textit{The} $\langle$\textit{name}$\rangle$ \textit{is a} $\langle$\textit{role}$\rangle$ \textit{of} $\langle$\textit{verb}$\rangle$'', such as ``\textit{The man is a agent of riding}''. For each object, we rank argument roles including \textsc{Other}, similarly to Multimedia Event Extraction. Following \cite{pratt2020grounded}, we ignore the \textsc{Place} argument role since it always not appear in the images.
The supervised setting is the same as Multimedia Event Extraction.

\textbf{Evaluation Metrics.} We follow~\cite{pratt2020grounded} to evaluate the accuracy of verb prediction (\textit{verb}), argument name prediction (\textit{value} for each argument and \textit{value-all} for all arguments of an event), and argument bounding box and name prediction (\textit{ground} for each argument and \textit{ground-all} for all arguments).

\subsection{VCR Implementation Details}

\textbf{Task Setting.}
VCR is a question answering task\footnote{\url{https://visualcommonsense.com/}}, including (1) \textit{Answer Prediction} from four options, and (2) \textit{Rationale Prediction} from four options to support the aforementioned answer. 

\textbf{Evaluation Goal.}
We include this task to evaluate whether event understanding can better support downstream tasks. 
To evaluate the quality of pretraining models, we adopt zero-shot settings solely relying on image-text alignment for a fair comparison. 

\textbf{Implementation.}
For Answer Prediction, we rank answers concatenated with questions.
For Rationale Prediction, we rank rationales by concatenating the question, the answer and the rationale. The ranking is based on both image alignment $d(i,t)$ and event graph alignment $d(G_{i}, G_{t})$. 
We also consider the question as query and concatenate them with the answer during ranking.

\subsection{VisualCOMET Implementation Details}
\label{app:visualcomet}

\textbf{Task Setting.}
Given the image and the event happening in the image with its participants, VisualCOMET~\cite{park2020visualcomet} aims to generate ``{intent}s'' showing what the participants ``\textit{need to do}'' %
before the image event, ``\textit{want to do}'' during the image event, and ``\textit{will most likely to do}'' after the image event. 

\textbf{Goal.} 
It necessitates a deep grasp of events and their connections, as well as a thorough comprehension of arguments roles. 

\begin{figure}[!h]
  \centering
   \includegraphics[width=0.8\linewidth,height=10em]{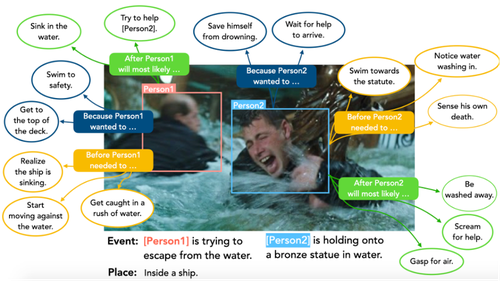}
   \caption{An example from VisualCOMET~\cite{park2020visualcomet}. }
   \label{fig:example-visualcomet}
\end{figure}

\textbf{Implementation.}
The input of VisualCOMET\footnote{ \url{https://visualcomet.xyz/}} is an image with events and participants, as shown in \cref{fig:example-visualcomet}. The output are intents, which is a short description of an event, such as ``\textit{swim to safety}'', ``\textit{sink in the water}'', etc. 
For each image and participant, we use intents from the training data as candidate intents, and rank them based on both image alignment $d(i,t)$ and event graph alignment $d(G_{i}, G_{t})$. 
The text is the concatenation of (1) input event description, (2) a temporal description (including ``\textit{before person1 need to}'', ``\textit{because person1 need to}'' and ``\textit{after person1 will most likely to}''), and (3) the candidate intents. 
For example, given the image with the input event ``\textit{person1 is trying to escape from the water}'', we concatenate it with the temporal description ``\textit{because person1 wanted to}'' and the candidate intent ``\textit{swim to safety while}''.
The ranking is based on both image alignment $d(i,t)$ and event graph alignment $d(G_{i}, G_{t})$, similar to Visual Commonsense Reasoning. 

\section{Effect of Text Information Extraction Performance}
Since text information extraction may have errors, we analyze its performance in the following sections.

\subsection{Text Event Extraction Performance Table}
The extraction performance of each component is shown in \cref{table:eval_result_component}, which achieves 72.1\% F-score on event extraction. 
\begin{table}[!htb]
\footnotesize
\centering
\setlength\tabcolsep{2pt}
\setlength\extrarowheight{1pt}
\begin{tabular}{c c c c c c} %
\toprule
\multicolumn{3}{l}{\textbf{Component}} & \textbf{Benchmark} & \textbf{Metric} & \textbf{Score} 
\\\midrule

\multicolumn{1}{l}{\multirow{3}{*}{\shortstack[l]{Event\\Mention\\Extraction}}} & \multirow{3}{*}{} & Entity & ACE+ERE & F\textsubscript{1} & 90.2 
\\ 
& & Trigger & ACE+ERE & F\textsubscript{1} & 72.8 
\\
&  & Argument & ACE+ERE & F\textsubscript{1} & 54.8 \\
&  & Relation & ACE+ERE & F\textsubscript{1} & 49.5

\\\midrule
\multicolumn{3}{l}{\multirow{2}{*}{\shortstack[l]{Document-level\\Argument Extraction}}} & ACE & {F\textsubscript{1}} & 66.7\\
& & & RAMS & F\textsubscript{1} & 48.6 
\\\midrule
\multirow{3}{*}{\shortstack[l]{Coreference\\Resolution}} & \multirow{2}{*}{} & Entity & OntoNotes & CoNLL & 92.4
\\
&  & Event & ACE & CoNLL & 84.8
\\
&  & Event & ERE-ES & CoNLL & 81.0
\\\bottomrule
\end{tabular}
\caption{Performance (\%) of each component. %
}
\label{table:eval_result_component}
\end{table}

\subsection{Event Type distribution}

\begin{figure}[!h]
  \centering
   \includegraphics[width=1.0\linewidth,height=10em]{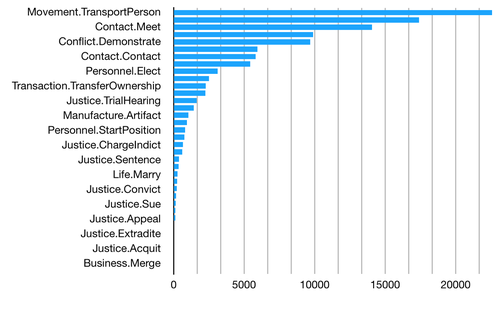}
   \caption{The top frequent event types from the event extraction results on VOANews captions. }
   \label{fig:voa_event_distribution}
\end{figure}

\begin{table*}[!hbt]
\footnotesize
\centering
\setlength\tabcolsep{2pt}
\setlength\extrarowheight{2pt}
\begin{tabular}{ l  | ccc|ccc | ccc|ccc | ccc|ccc}
\toprule
\multirow{1}{*}{\textbf{Model}} & \multicolumn{6}{c|}{\textbf{Flickr30k}} & \multicolumn{6}{c|}{\textbf{MSCOCO}} & \multicolumn{6}{c}{\textbf{VOANews}}
\\
 & \multicolumn{3}{c|}{text-to-image} & \multicolumn{3}{c|}{image-to-text} & \multicolumn{3}{c|}{text-to-image} & \multicolumn{3}{c|}{image-to-text} &
 \multicolumn{3}{c|}{text-to-image} &
 \multicolumn{3}{c}{image-to-text} 
\\
& R@1 & R@5 & R@10 & R@1 & R@5 & R@10 & R@1 & R@5 & R@10 & R@1 & R@5 & R@10 & R@1 & R@5 & R@10 & R@1 & R@5 & R@10
\\
\midrule
{CLIP}
& 62.2 & 85.9 & 91.7 & 81.9 & 95.0 & 97.5 & 30.3 & 55.0 & 66.4 & 50.3 & 75.7 & 84.0 
& 21.2 %
& 63.4 & 74.7
& 23.4 %
& 63.1 & 73.9
\\
{CLIP} pretrained on news
& 64.3 & 87.5 & 92.7 & 81.2 & 95.4 & 98.2 & 32.2 & 57.4 & 68.4 & 50.8 & 75.6 & 83.8 
& 23.5  %
& 69.5 & 79.9
& 25.1  %
& 70.2 & 80.1
\\
\midrule
\textbf{CLIP-Event}
& \textbf{67.0} &\textbf{89.0} & \textbf{93.9} & \textbf{82.6} & \textbf{95.9} & \textbf{98.4} & \textbf{34.0} & \textbf{59.4} & \textbf{70.5} & \textbf{51.3} & \textbf{76.0} & \textbf{84.0} 
& \textbf{27.5}  %
& \textbf{70.7} & \textbf{82.1}
& \textbf{28.7}  %
& \textbf{71.0} & \textbf{81.0}
\\
\ \ \  w/o OptimalTransport
& 65.6 & 88.3 & 93.6 & 80.5 & 94.8 & 97.4 & 32.5 & 58.0 & 68.9 & 51.0 &75.2 & 82.9  
& 25.5 %
& 70.6 & 80.7
& 26.9 %
& 70.4 & 80.5
\\
\bottomrule
\end{tabular}
\vspace{-.1cm}
\caption{
R@1(\%), R@5(\%), R@10(\%) on image retrieval on Flickr30k (1k test), MSCOCO (5k test) and VOANews. %
}
\label{table:result_retrieval_large}
\end{table*}

As shown in \cref{fig:voa_event_distribution}, the events extracted from captions are primarily visually detectable events, i.e., the. events can be depicted in the images.

{\small
\bibliographystyle{ieee_fullname}
\bibliography{ref}
}

\end{document}